\documentclass[10pt]{article}
\usepackage{amsmath,amsfonts,bm}

\def\eqref#1{equation~\ref{#1}}
\def\1{\bm{1}}

\def\vmu{{\bm{\mu}}}

\def\vv{{\bm{v}}}
\def\vw{{\bm{w}}}
\def\vx{{\bm{x}}}
\def\vy{{\bm{y}}}
\def\vz{{\bm{z}}}

\def\mW{{\bm{W}}}

\DeclareMathAlphabet{\mathsfit}{\encodingdefault}{\sfdefault}{m}{sl}
\SetMathAlphabet{\mathsfit}{bold}{\encodingdefault}{\sfdefault}{bx}{n}

\DeclareMathOperator*{\argmax}{arg\,max}
\DeclareMathOperator*{\argmin}{arg\,min}

\usepackage[utf8]{inputenc} 
\usepackage[T1]{fontenc}    
\usepackage[pdftex]{graphicx}
\usepackage{caption}  
\usepackage[sort,compress]{natbib}
\usepackage{hyperref}     
\usepackage{url}          
\usepackage{booktabs}      
\usepackage{amsfonts}      
\usepackage{nicefrac}      
\usepackage{microtype}     
\usepackage{xcolor}       
\usepackage{wrapfig}
\usepackage{subcaption}
\usepackage{tabularx}
\usepackage{amsmath}
\usepackage{amssymb}
\usepackage{algorithm}
\usepackage{algpseudocode}
\usepackage{mathtools}
\usepackage{copyrightbox}
\usepackage{bbm}
\usepackage{amsthm} 
\usepackage{chngcntr}
\usepackage{enumitem}
\usepackage{xspace}
\usepackage{placeins}
\usepackage{titlesec}
\usepackage{graphicx}
\usepackage{arxiv}
\usepackage{doi}
\usepackage{changes}

\DeclareCaptionLabelSeparator{pipe}{\textbf{ | }}
\DeclareCaptionLabelFormat{custom}{\textbf{Fig. #2}}
\title{Dimensions underlying the representational alignment of deep neural networks with humans}

\author{\large \normalfont Florian P. Mahner \textsuperscript{1,2,\thanks{Equal contribution}}\ , Lukas Muttenthaler \textsuperscript{1,3,4,\footnotemark[1]\hspace{0.4em}\thanks{Work partly done while a Student Researcher at Google DeepMind}}\ ,
 Umut G\"u\c{c}l\"u \textsuperscript{2}, Martin N. Hebart \textsuperscript{1,5,6} \\
  \\ 
  \textsuperscript{1} Max Planck Institute for Human Cognitive and Brain Sciences, Leipzig, Germany \\
  \textsuperscript{2} Donders Institute for Brain, Cognition and Behaviour, Nijmegen, the Netherlands \\
  \textsuperscript{3} Machine Learning Group, Technische Universit\"at Berlin, Germany \\
  \textsuperscript{4} Berlin Institute for the Foundations of Learning and Data (BIFOLD), Berlin, Germany \\
  \textsuperscript{5} Department of Medicine, Justus Liebig University, Giessen, Germany \\
  \textsuperscript{6} Center for Mind, Brain and Behavior, Universities of Marburg, Giessen, and Darmstadt, Germany}

\date{} 
\renewcommand{\headeright}{}
\renewcommand{\undertitle}{}
\renewcommand{\shorttitle}{}

\hypersetup{
pdftitle={Dimensions underlying the representational alignment of deep neural networks with humans},
pdfauthor={Florian P. Mahner, Lukas Muttenthaler, et al.},
}

\begin{document}
\begin{NoHyper}
\maketitle
\end{NoHyper}

%TC:ignore
\begin{abstract}
Determining the similarities and differences between humans and artificial intelligence (AI) is an important goal both in computational cognitive neuroscience and machine learning, promising a deeper understanding of human cognition and safer, more reliable AI systems. Much previous work comparing representations in humans and AI has relied on global, scalar measures to quantify their alignment. However, without explicit hypotheses, these measures only inform us about the degree of alignment, not the factors that determine it. To address this challenge, we propose a generic framework to compare human and AI representations, based on identifying latent representational dimensions underlying the same behavior in both domains. Applying this framework to humans and a deep neural network (DNN) model of natural images revealed a low-dimensional DNN embedding of both visual and semantic dimensions. In contrast to humans, DNNs exhibited a clear dominance of visual over semantic properties, indicating divergent strategies for representing images. While in-silico experiments showed seemingly consistent interpretability of DNN dimensions, a direct comparison between human and DNN representations revealed substantial differences in how they process images. By making representations directly comparable, our results reveal important challenges for representational alignment and offer a means for improving their comparability.
\end{abstract}
%TC:endignore

\section*{} 

Deep neural networks (DNNs) have achieved impressive performance, matching or surpassing human performance in various perceptual and cognitive benchmarks, including image classification \citep{krizhevsky2012imagenet, he2016deep}, speech recognition \citep{hinton2012deep, amodei2016deep} and strategic gameplay \citep{silver2016mastering, vinyals2019grandmaster}. In addition to their excellent performance as machine learning models, DNNs have drawn attention in the field of computational cognitive neuroscience for their notable parallels to cognitive and neural systems in humans and animal models \citep{khaligh2014deep, yamins2014performance, gucclu2015deep, rajalingham2015comparison, kubilius2016deep}. These similarities, observed through different types of behavior or patterns of brain activity, have sparked a growing interest in determining both the factors underlying these similarities and the differences between human and DNN representations. From the machine learning perspective, understanding where DNNs exhibit a limited alignment with humans can support the development of better and more robust artificial intelligence systems. From the perspective of computational cognitive neuroscience, DNNs with stronger human alignment promise to be better candidate computational models of human cognition and behavior \citep{cichy2019deep, lindsay2021convolutional, kanwisher2023using, doerig2023neuroconnectionist}. 

Much previous research on the alignment of human and artificial visual systems has compared behavioral strategies (e.g., classification) in both systems and has revealed important limitations in the generalization performance of DNNs \citep{rajalingham2018large, geirhos2018generalisation, rosenfeld2018elephant, beery2018recognition, szegedy2013intriguing}. Other work has focused on directly comparing cognitive and neural representations in humans to those in DNNs, using methods such as representational similarity analysis \citep[RSA;][]{kriegeskorte2008representational} or linear regression \citep{attarian2020transforming, roads2020learning, peterson2018evaluating, muttenthaler2022human}. This quantification of alignment has led to a direct comparison of numerous DNNs across diverse visual tasks \citep{conwell2022can, schrimpf2018brain, muttenthaler2023improving, wang2023better}, highlighting the role of factors such as architecture, training data, or learning objective in determining the similarity to humans \citep{storrs2021diverse, conwell2022can, muttenthaler2022human, wang2023better}.

Despite the appeal of summary statistics, such as correlation coefficients or explained variance, for comparing the representational alignment of DNNs with humans, they only quantify the \textit{degree} of representational or behavioral alignment. However, without explicit hypotheses about potential causes for misalignment, such scalar measures are limited in their explanatory scope of which properties determine this degree of alignment, that is, which representational factors underlie the similarities and differences between humans and DNNs. While diverse methods for interpreting DNN activations have been developed at various levels of analysis, ranging from single units to entire layers \citep{erhan2009visualizing, zeiler2014visualizing, zhou2018revisiting, morcos2018importance, bau2020understanding}, the direct comparability to human representations has remained a key challenge.  

Inspired by recent work in the cognitive sciences that has revealed core visual and semantic representational dimensions underlying human similarity judgments of object images \citep{hebart2020revealing}, here we propose a framework to systematically analyze and compare the dimensions that shape representations in humans and DNNs. In this work, we apply this framework to human visual similarity judgments and representations in a DNN trained to classify natural images. Our approach reveals numerous interpretable DNN dimensions that appear to reflect both visual and semantic image properties and that appear to be well-aligned to humans. In contrast to humans, who showed a dominance of semantic over visual dimensions, DNNs exhibited a striking visual bias, demonstrating that downstream semantic behavior is driven more strongly by different, primarily visual, strategies. While psychophysical experiments on DNN dimensions underscored their global interpretability, a direct comparison with human dimensions revealed that DNN representations in fact only approximate human representations but lack the consistency expected from property-specific visual and semantic dimensions. Together, our results reveal key factors underlying the representational alignment and misalignment between humans and DNNs, shed light on potentially divergent representational strategies, and highlight the potential of this approach to identify the factors underlying the similarities and differences between humans and DNNs.

\section{Results}

To improve the comparability of human and DNN representations, we aimed to identify the similarities and differences in core dimensions underlying {human and DNN representations of images}. To achieve this aim, we treated the neural network analogously to a human participant carrying out a cognitive behavioral experiment and then derived representational embeddings using a recent variational embedding technique \citep{muttenthaler2022vice} both from human similarity judgments and a DNN on the same behavioral task (see Methods). This approach ensured direct comparability between human and DNN representations. As a behavioral task, we chose a triplet odd-one-out similarity task, where from a set of three object images $i$, $j$, $k$, participants have to select the most dissimilar object (Fig. \ref{fig:overview}a; see Supplementary Information \ref{sec:task-validation} for an analysis of the role of task instructions on triplet choice behavior). In this task, the perceived similarity between two images $i$ and $j$ is defined as the probability of choosing these images to belong together across varying contexts imposed by a third object image $k$. By virtue of providing minimal contexts, the odd-one-out task highlights the information sufficient to capture the similarity between object images $i$ and $j$ across diverse contexts. In addition, it approximates human categorization behavior for arbitrary visual and semantic categories, even for quite diverse sets of objects \citep{zheng2019revealing,hebart2020revealing, muttenthaler2022vice}. Thus, by focusing on the building blocks of categorization which underlies diverse behaviors, this task is ideally suited for comparing object representations between humans and DNNs.

\begin{figure}[!t]
    \centering
    \includegraphics[width=\textwidth]{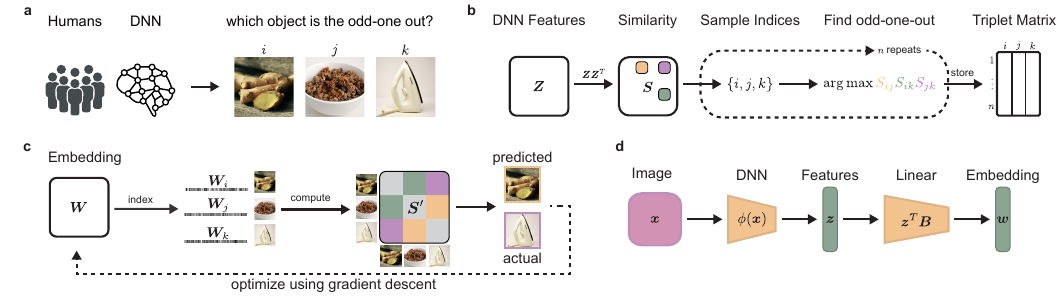}
    \caption{\textbf{Overview: A computational framework that captures core DNN object representations in analogy to humans by simulating behavioral decisions in an odd-one-out task.} \textbf{a}, The triplet odd-one-out task, where a human participant or a DNN is presented a set of three images and is asked to select the image that is most different from the others. \textbf{b}, Sampling approach of odd-one-out decisions from DNN representations. First, a dot-product similarity space is constructed from DNN features. Next, for a given triplet of objects, the most similar pair in this similarity space is identified, making the remaining object the odd-one-out. For humans, this sampling approach is based on observed behavior, which is used as a measure of their internal cognitive representations. \textbf{c}, Illustration of the computational modeling approach to learn a lower-dimensional object representation for human participants and the DNN, optimized to predict behavioral choices made in the triplet task. \textbf{d}, Schematic depiction of the interpretability pipeline that allows for the prediction of object embeddings from pretrained DNN features.}
    \label{fig:overview}
\end{figure}

\begin{figure}[!t]
    \centering
    \includegraphics[width=\textwidth]{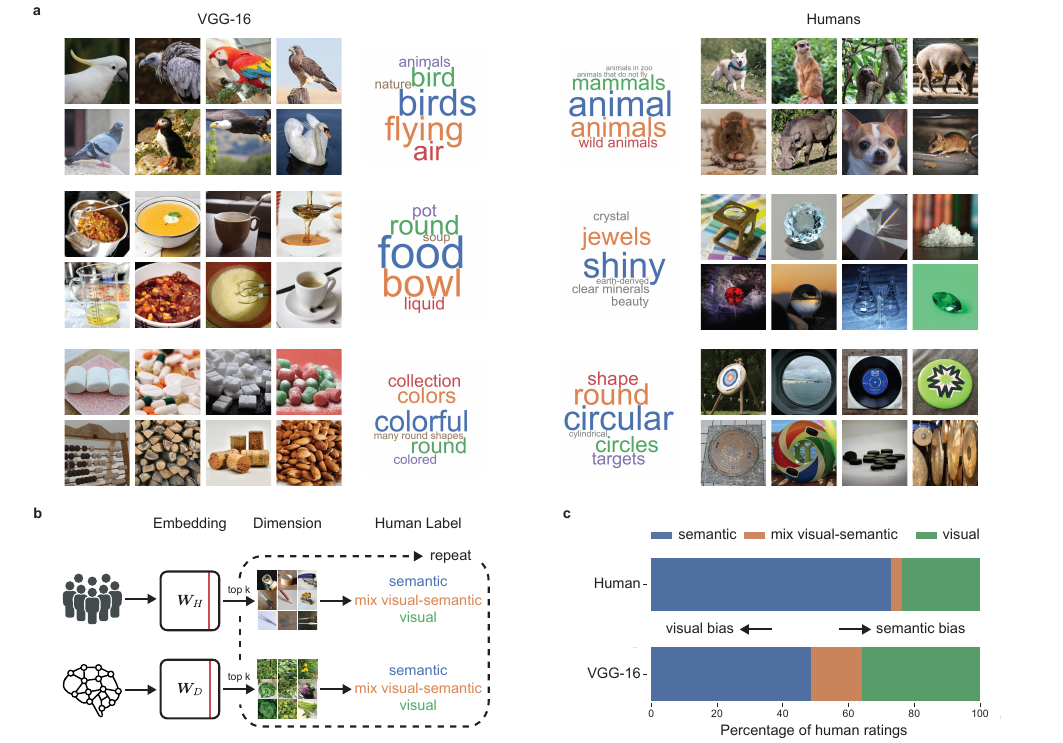}
    \caption{\textbf{Representational embeddings inferred from human and DNN behavior.} \textbf{a}, Visualization of example dimensions from human- and DNN-derived representational embeddings, with a selection of dimensions that had been rated as semantic, mixed visual-semantic, and visual, alongside their dimension labels obtained from human judgments. Note that the displayed images reflect only images with a public domain license and not the full image set \citep{stoinski2023thingsplus} \textbf{b}, Rating procedure for each dimension, which was based on visualizing the top $k$ images according to their numeric weights. Human participants labeled each of the human and DNN dimensions as predominantly semantic, visual, mixed visual-semantic, or unclear (unclear ratings not shown: 7.35\% of all dimensions for humans, 8.57\% for VGG-16). \textbf{c}, Relative importance of dimensions labeled as visual and semantic, where VGG-16 exhibited a dominance of visual and mixed dimensions relative to humans that showed a clear dominance of semantic dimensions.}
    \label{fig:dimensions}
\end{figure}

For human behavior, we used a set of 4.7 million publicly available odd-one-out judgments \citep{hebart2023thingsdata} over 1,854 diverse object images, derived from the THINGS object concept and image database \citep{hebart2019things}. For the DNN, we collected similarity judgments for 24,102 images of the same objects used for humans (1,854 objects, 13 examples per object). We used a larger set of object images since the DNN was less limited by constraints in dataset size than humans. This allowed us to obtain more precise estimates of their representation. To derive DNN representations, we chose a pretrained VGG-16 model \citep{simonyan2014very}, given its common use in the computational cognitive neurosciences. Specifically, this network has been shown to exhibit good correspondence to both human behavior \citep{geirhos2018generalisation} and measured neural activity \citep{gucclu2015deep, schrimpf2018brain, nonaka2021brain} and performs well at predicting human similarity judgments \citep{jozwik2017deep, peterson2018evaluating, king2019similarity, storrs2021diverse, muttenthaler2022human, kaniuth2024high}. VGG-16 was trained on the 1000-class ImageNet dataset \citep{deng2009imagenet}, which contains a diverse range of image categories, such as animals, everyday objects, and scenes. However, for completeness, we additionally ran similar analyses for a broader range of neural network architectures (see Supplementary Information \ref{sec:modelcomp}). We focused on the penultimate layer activations as they are closest to the behavioral output, and they also showed closest representational correspondence to humans (see Supplementary Information \ref{sec:layercomp}). For the DNN, we generated a dataset of behavioral odd-one-out choices for the 24,102 object images (Fig. \ref{fig:overview}b). To this end, we first extracted the DNN layer activations for all images. Next, for a given triplet of activations $\vz_i$, $\vz_j$ and $\vz_k$, we computed the dot product between each pair as a measure of similarity, then identified the most similar pair of images in this triplet, and designated the remaining third image as the odd-one-out. Given the excessively large number of possible triplets for all 24,102 images, we approximated the full set of object choices from a random subset of $20$ million triplets \citep{lalit2016finite}. 

From both sets of available triplet choices, we next generated two representational embeddings, one for humans and one for the DNN, where each embedding was optimized to predict the odd-one-out choices in humans and DNNs, respectively. In these embeddings, each object is described through a set of dimensions that represent interpretable object properties. To obtain these dimensions and for comparability to previous work in humans \citep{zheng2019revealing, hebart2020revealing, muttenthaler2022vice}, we imposed sparsity and non-negativity constraints on the optimization, which support their interpretability and provide cognitively plausible criteria for dimensions\citep{hoyer2002non, murphy2012learning, fyshe2015compositional, hebart2020revealing, hebart2023thingsdata}. Sparsity constrained the embedding to consist of fewer dimensions, motivated by the observation that real-world objects are typically characterized by only a few properties. Non-negativity encouraged a parts-based description, where dimensions cannot cancel each other out. Thus, a dimension’s weight indicated its relevance in predicting an object’s similarity to other objects. During training, each randomly initialized embedding was optimized using a recent variational embedding technique \citep{muttenthaler2022vice} (for details, see Methods: \nameref{sec:embedding-optim-and-pruning}). The optimization resulted in two stable, low-dimensional embeddings, with 70 reproducible dimensions for the DNN embedding and 68 for the human embedding. The DNN embedding captured 84.03\% of the total variance in image-to-image similarity, while the human embedding captured 82.85\% of the total variance and 91.20\% of the explainable variance given the empirical noise ceiling of the dataset.

\subsection{DNN dimensions reflect diverse image properties}

 Having identified stable, low-dimensional embeddings that are predictive of triplet odd-one-out judgments, we first assessed the interpretability of each identified DNN dimension by visualizing object images with large numeric weights. In addition to this qualitative assessment, we validated these observations for the DNN by asking 12 (6 female, 6 male) human participants to provide labels for each dimension separately (see Methods: \nameref{sec:labeling-dimensions}). Similar to the core semantic and visual dimensions underlying odd-one-out judgments in humans described previously \citep{hebart2020revealing, muttenthaler2022vice,hebart2023thingsdata}, the DNN embedding yielded many interpretable dimensions, which appeared to reflect both semantic and visual properties of objects. The semantic dimensions included taxonomic membership (e.g. food-related, technology-related, home-related) and other knowledge-related properties (e.g. softness), while the visual dimensions reflected visual-perceptual attributes (e.g. round, green, stringy), with some dimensions reflecting a composite of semantic and visual properties (e.g. green and organic) (Fig. \ref{fig:dimensions}a). Of note, the DNN dimensions also revealed a sensitivity to basic shapes, including roundness, boxiness and tube-shape. This suggests that, in line with earlier studies \citep{hermann2020origins, singer2022photos}, DNNs indeed learn to represent basic shape properties, an aspect that might not be apparent in their overt behavior \citep{geirhos2019imagenet}.
 
 Despite the apparent similarities, there were, however, also striking differences found between humans and the DNN. First, overall, DNN dimensions were less interpretable than human dimensions, as confirmed by the evaluation of all dimensions by two independent raters (see Supplementary Information \ref{sec:interpretability}). This indicates a global difference in how the DNN assigns images as being conceptually similar to each other. Second, while human dimensions were clearly dominated by semantic properties, many DNN dimensions were more visual-perceptual in nature or reflected a mixture of visual and semantic information. We quantified this observation by asking the same two independent experts to rate human and DNN dimensions according to whether they were primarily visual-perceptual, semantic, reflected a mixture of both, or were unclear (Fig. \ref{fig:dimensions}b). To confirm that the results were not an arbitrary byproduct of the chosen DNN architecture, we provided the raters with four additional DNNs for which we had computed additional representational embeddings. The results revealed a clear dominance of semantic dimensions in humans, with only a small number of mixed dimensions. In contrast, for DNNs we found a consistently larger proportion of dimensions that were dominated by visual information or that reflected a mixture of both visual and semantic information (Fig. \ref{fig:dimensions}c,  Supplementary Fig. \ref{fig:rsa-model-comps}b for all DNNs). This visual bias is also present across intermediate representations of VGG-16 and even stronger in early to late convolutional layers (see Supplementary Fig. \ref{fig:early-late}). This demonstrates a clear difference in the relative weight that humans and DNNs assign to visual and semantic information, respectively. We independently validated these findings using a semantic text embedding and observed a similar pattern of visual bias (see Supplementary Information \ref{sec:sem-validation}, indicating that the results were not solely a product of human rater bias).

\subsection{Linking DNN dimensions to their interpretability}

\begin{figure}[!t]
    \centering
    \includegraphics[width=\textwidth]{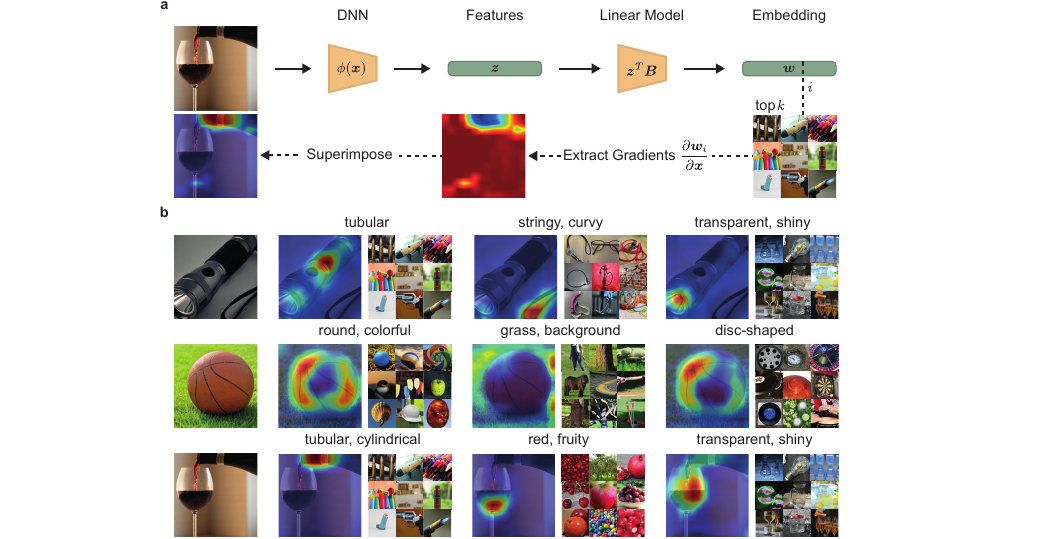}
    \caption{\textbf{Relevance of image properties for embedding dimension.} \textbf{a}, General methodology of the approach. We used Grad-CAM \citep{selvaraju2017grad} to visualize the importance of distinct image parts based on the gradients of the penultimate DNN features that we initially used to sample triplet choices. The gradients were obtained in our fully differentiable interpretability model with respect to a dimension $\vw$ in our embedding. \textbf{b}, We visualize the heatmaps for three different images and dimensions. Each column shows the relevance of parts of an image for that dimension. For this figure, we filtered the embedding by images available in the public domain \citep{stoinski2023thingsplus}, except for two images sourced from Flickr under a CC BY 2.0 license: the flashlight by \citet{cborysiuk} and the wineglass by \citet{wszkutnik}.}
    \label{fig:relevance}
\end{figure}

Despite the overall differences in human and DNN representational dimensions, the DNN also contained many dimensions that appeared to be interpretable and comparable to those found in humans. Next, we aimed at testing to what degree these interpretable dimensions truly reflected specific visual or semantic properties, or whether they only superficially appeared to show this correspondence. To this end, we experimentally and causally manipulated images and observed the impact on dimension scores. Beyond general interpretability, these analyses further establish which visual properties in each image drive individual dimensions and thus determine image representations. 

Image manipulation requires a direct mapping from input images to the embedding dimensions. However, the embedding dimensions were derived using a sampling-based optimization based on odd-one-out choices inferred from penultimate DNN features. Consequently, this approach does not directly map these features to the learned embedding. To establish this mapping, we applied $\ell_{2}$-regularized linear regression to link the DNN’s penultimate layer activations to the learned embedding. This mapping then enables the prediction of embedding dimensions from the penultimate features activations in response to novel or manipulated images. (Fig. \ref{fig:overview}d). Penultimate layer activations were indeed highly predictive of each embedding dimension, with all dimensions exceeding an $R^2$ of 75\%, and the majority exceeding 85\%. Thus, this allowed us to accurately predict dimension values for novel images.

Having established an end-to-end mapping between input image and individual object dimensions, we next used three approaches to both probe the consistency of the interpretation and identify dimension-specific image properties. First, to identify image regions relevant for each individual dimension, we used Grad-CAM \citep{selvaraju2017grad}, an established technique for providing visual explanations. Grad-CAM generates heatmaps that highlight the image regions that are most influential for model predictions. Unlike the typical use of Grad-CAM, which focuses on generating visual explanations for model classifications (e.g., dog vs. cat), we employed Grad-CAM to reveal which image regions drive the dimensions in the DNN embedding. The results of this analysis are illustrated with example images in Fig. \ref{fig:relevance}. Object dimensions were indeed driven by different image regions that contain relevant information, in line with the dimension's interpretation derived from human ratings and suggesting that the representations captured by the DNN's penultimate layer allow us to distinguish between different parts of the image that carry different functional significance. 

\begin{figure}[!t]
    \centering
    \includegraphics[width=\textwidth]{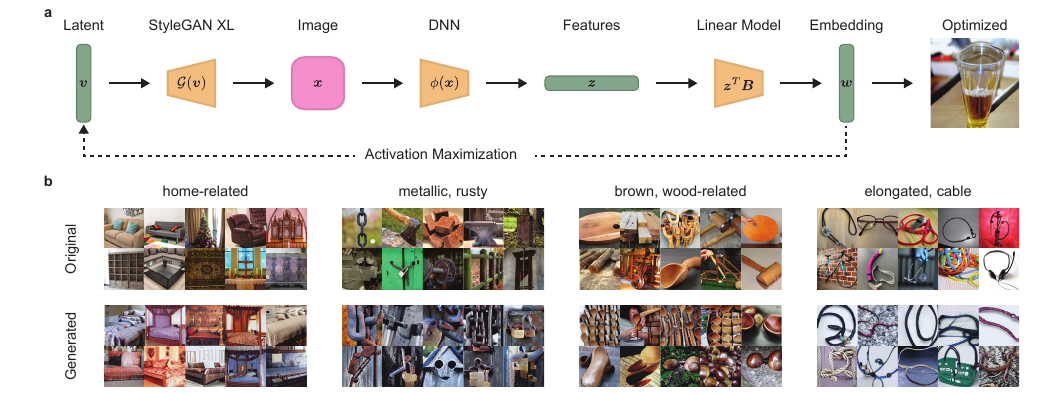}
    \caption{\textbf{Maximally activating images for embedding dimensions.} \textbf{a}, Using StyleGAN XL \citep{sauer2022stylegan}, we optimized a latent code to maximize the predicted response in a specific embedding dimension. \textbf{b}, Visualizations for different dimensions in our embedding. We show the top $10$ images that score highest in the dimension and the corresponding top $10$ generated images.  For this figure, we filtered the embedding by images available in the public domain \citep{stoinski2023thingsplus}}
    \label{fig:act-max}
\end{figure}

As a second image explanation approach, to highlight which image properties drive a dimension, we used a generative image model to create novel images optimized for maximizing values of a given dimension \citep{erhan2009visualizing, yosinski2015understanding, montavon2018methods}. Unlike conventional activation maximization targeting a single DNN unit or a cluster of units, our approach aimed to selectively amplify activation in dimensions of the DNN embedding across the entire DNN layer, using a pretrained generative adversarial neural network \citep[StyleGAN XL;][]{sauer2022stylegan}). To achieve this, we applied our linear end-to-end mapping to predict the embedding dimensions from the penultimate activations in response to the images generated by StyleGAN-XL. The results of this procedure are shown in  Figure \ref{fig:act-max}b. The approach successfully generated images with high numerical values in the dimensions of our DNN embedding. Indeed, the properties highlighted by these generated images appear to align with human assigned labels for each specific dimension, again suggesting that the DNN embedding contained conceptually meaningful and coherent object properties similar to those found in humans.

As a third image explanation approach, given that different visual properties naturally co-occur across images, and to tease apart their respective contribution, we causally manipulated individual image properties and observed the effect on predicted DNN dimensions. We exemplify this approach with manipulations in color, object shape, and background (see Supplementary Information \ref{sec:causal}), largely confirming our predictions, showing specific activation decreases or increases in dimensions that appeared to be representing these properties.

\subsection{Factors driving human and DNN similarities and differences}

\begin{figure}[!t]
    \centering
    \includegraphics[width=\textwidth]{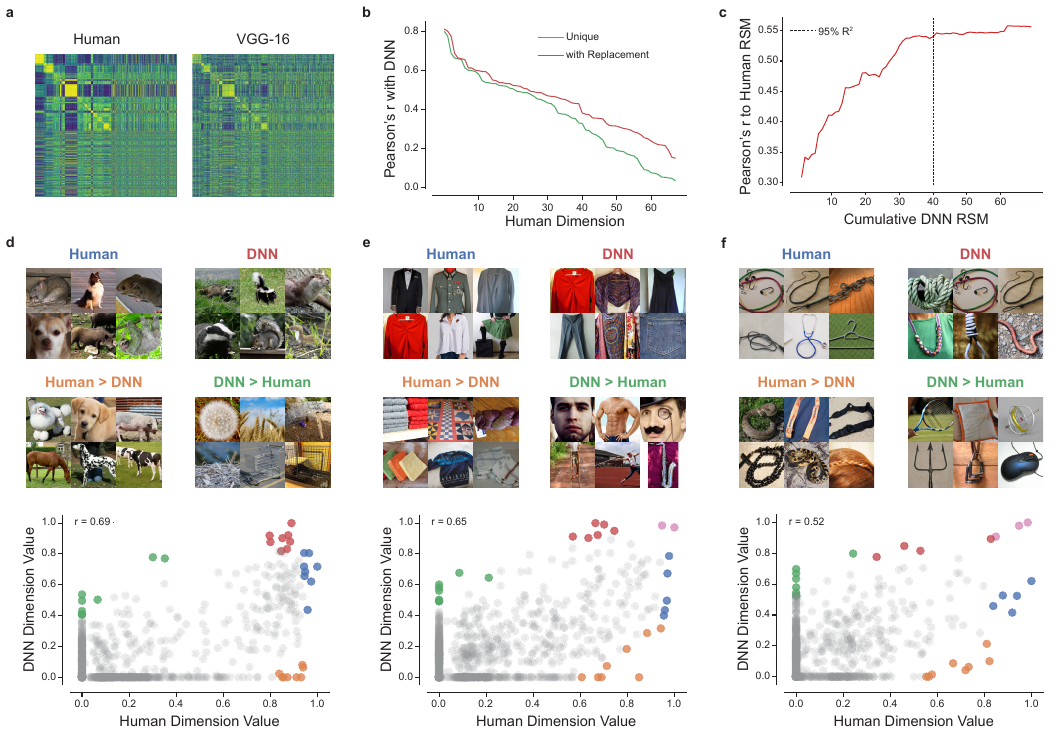}
    \caption{\textbf{Factors that determine the similarity between human and VGG-16 embedding dimensions.} \textbf{a}, Representational similarity matrices reconstructed from the human and VGG-16 embedding. Each row represents an object, with rows sorted into 27 superordinate categories (e.g., animal, food, furniture) from \citep{hebart2019things} to better highlight similarities and differences in representation. \textbf{b}, Pairwise correlations between human and VGG-16 embedding dimensions. \textbf{c}, Cumulative RSA analysis that shows the amount of variance explained in the human RSM as a function of the number of DNN dimensions. The black line shows the number of dimensions required to explain 95\% of the variance. \textbf{d-f}, \textit{Intersection} (red and blue regions) and \textit{differences} (orange and green regions) between three highly correlating human and DNN dimensions. Pink circles denote the intersection of the red and blue regions, i.e., where the same image scores highly in both dimensions. For this figure, we filtered the embedding by images from the public domain.}
    \label{fig:factors}
\end{figure}

\begin{figure}[!t]
    \centering
      \includegraphics[width=1\textwidth]{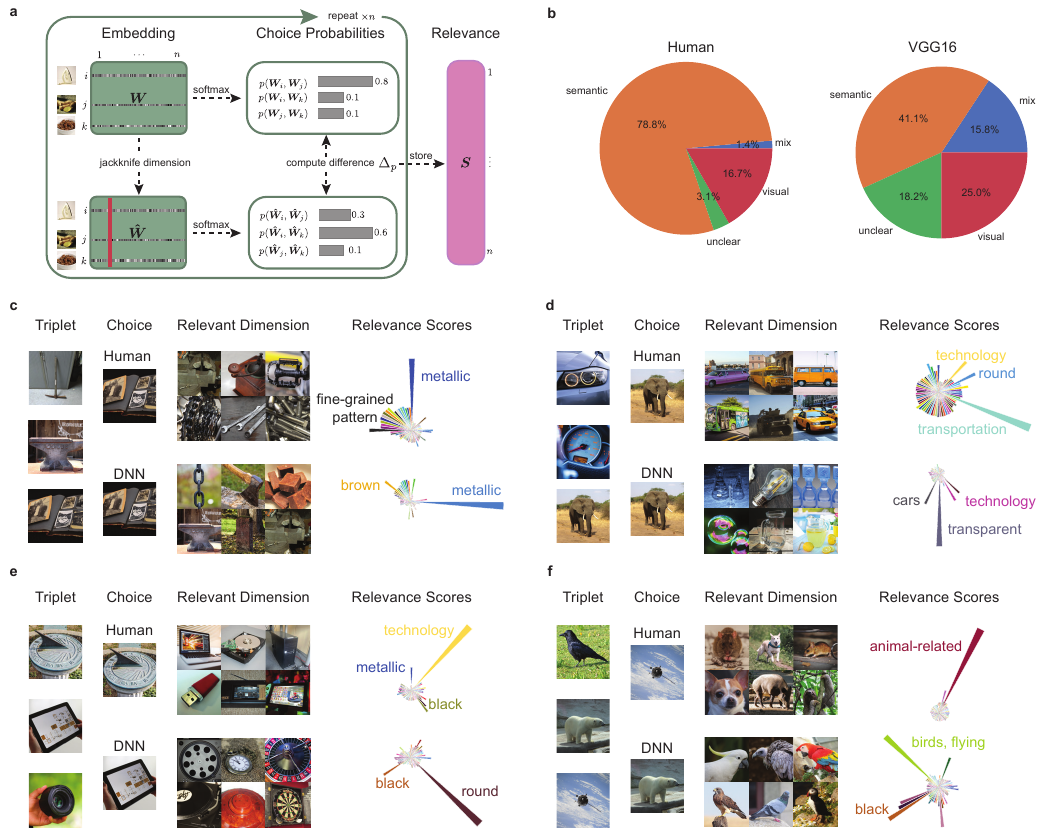}
    \caption{\textbf{Overt behavioral choices in humans and the DNN.} \textbf{a}, Overview of the approach. For one triplet, we computed the original predicted softmax probability based on the entire representational embedding for each object image in the triplet. We then iteratively pruned individual dimensions from the representational embedding and stored the difference to the predicted difference to the original softmax probability of the entire embedding as a relevance score for that dimension. \textbf{b}, We calculated the relevance scores for a random sample of 10 million triplets and identified the most relevant dimension for each triplet. We then labeled the 10 million most relevant dimensions according to human-labeled visual properties as semantic, mixed visual-semantic, visual or unclear. Semantic dimensions are most relevant for human behavioral choices, whereas for VGG-16 visual and mixed semantic-visual properties are more relevant. \textbf{c-f}, We rank sorted changes in softmax probability to find triplets where human and the DNN maximally diverge. Each panel shows a triplet with the behavioral choice made by humans and the DNN. We visualized the most relevant dimension for that triplet alongside the distribution of relevance scores. Each dimension is assigned its human annotated label. For this figure, we filtered the embedding by images from the public domain.}
    \label{fig:jackknife}
\end{figure}

The previous results have confirmed the overall consistency and interpretation of the DNN's visual and semantic dimensions based on common interpretability techniques. However, a direct comparison with human image representations is crucial for identifying which representational dimensions align well and which do not. Traditional representational similarity analysis (RSA) provides a global metric of representational alignment, revealing a moderate correlation (Pearson's $r = 0.55$) between the representational similarity matrices (RSMs) of humans and the DNN (Fig. \ref{fig:factors}a). While this indicates some degree of alignment in object image representations, it does not clarify the factors driving this alignment. To address this challenge, we directly compared pairs of dimensions from both embeddings, pinpointing which dimensions contributed the most to the overall alignment and which dimensions were less well aligned.

For each human dimension, we identified the most strongly correlated DNN dimension, once without replacement (unique) and once with replacement, and sorted the dimensions based on their fit (Fig. \ref{fig:factors}b). This revealed a close alignment, with Pearson’s reaching up to $r=0.80$ for a select few dimensions which gradually declined across other representational dimensions. To determine whether the global representational similarity was driven by just a few well-aligned dimensions or required a broader spectrum of dimensions, we assessed the number of dimensions needed to explain human similarity judgments. The analysis revealed that 40 dimensions were required to capture 95\% of the variance in representational similarity with the human RSM (Fig. \ref{fig:factors}c). Although this number is much smaller than the original 4096-dimensional VGG-16 layer, these results demonstrate that the global representational similarity is not solely driven by a small number of well-aligned dimensions.

Given the imperfect alignment of DNN and human dimensions, we explored the similarities and differences in the stimuli represented by these dimensions. For each dimension, we identified which images were most representative of both humans and the DNN. Crucially, to highlight the discrepancies between the two domains, we then identified which images exhibited strong dimension values for humans but weak values for the DNN, and vice versa (Fig. \ref{fig:factors}d-f). While the results indicated similar visual and semantic representations in the most representative images, they also exposed clear divergences in dimension meanings. For instance, in an animal-related dimension, humans consistently represented animals even for images where the DNN exhibited very low dimension values. Conversely, the DNN dimension strongly represented objects that were not animals, such as natural objects, cages, or mesh (Fig. \ref{fig:factors}d). Similarly, a string-related dimension maintained a string-like representation in humans but included other objects in the DNN that were not string-like, potentially reflecting properties related to thin, curvy objects or specific image properties (Fig. \ref{fig:factors}f).

\subsection{Relevance of object dimensions for categorization behavior}

Since internal representations do not necessarily translate into behavior, we next addressed whether this misalignment would translate to downstream behavioral choices. To this end, we employed a jackknife resampling procedure to determine the relevance of individual dimensions for odd-one-out choices. For each triplet, we iteratively pruned dimensions in both the human and DNN embeddings and observed changes in the predicted probabilities of selecting the odd-one-out, yielding an importance score for each dimension for the odd-one-out choice (Fig. \ref{fig:jackknife}a). 
The results of this analysis showed that while humans and DNNs often aligned in both their representations and choices, a sizable fraction of choices exhibited the same behavior despite strong differences in representations (Fig. \ref{fig:jackknife}b). For behavioral choices, the semantic bias in humans was enhanced, as evidenced by an even stronger importance of semantic relative to visual or mixed dimensions in humans as compared to DNNs. Individual triplet choices were affected not only by semantic but also by visual dimensions (Fig. \ref{fig:jackknife}c-f). Together, these results demonstrate that the differences in how humans and DNNs represent object images not only translate into behavioral choices but are also further amplified in their categorization behavior. 

\section{Discussion}

A key challenge in understanding the similarities and differences in humans and artificial intelligence (AI) lies in establishing ways to make these two domains directly comparable. Overcoming this challenge would allow us to identify strategies to make AI more human-like \citep{geirhos2018generalisation} and for using AI as effective models of human perception and cognition. In this work, we propose a framework to identify interpretable factors that determine the similarities and differences between human and AI representations. In this framework, these factors can be identified by using the same experiment to probe behavior in humans and AI systems and applying the same computational strategy to the natural and artificial responses to infer their respective interpretable embeddings. We applied this approach to human similarity judgments and representations of DNNs trained on natural images with varying objectives, with a primary focus on an image classification model. This allowed for a direct, meaningful comparison of the representations underlying human similarity judgments with the representations of the image classification model.

Our results revealed that the DNN contained representations that appeared to be similar to those found in humans, ranging from visual (e.g., "white", "circular/round", "transparent") to semantic properties (e.g., "food-related", "fire-related"). However, a direct comparison to humans showed largely different strategies for arriving at these representations. While human representations were dominated by semantic dimensions, the DNN exhibited a pronounced bias towards visual or mixed visual-semantic dimensions. In addition, a direct comparison of seemingly aligned dimensions revealed that DNNs only approximated the semantic representations found in humans. These different strategies were also reflected in their behavior, where similar behavioral outcomes were based on different embedding dimensions. Thus, despite seemingly well aligned human and DNN representations at a global level, deriving dimensions underlying the representational similarities provided a more complete and more fine-grained picture of this alignment, revealing the nature of the representational strategies that humans and DNNs use \citep{cichy2019deep, sucholutsky2023getting, kanwisher2023using}.

While approaches like RSA \citep{kriegeskorte2008representational, kornblith2019similarity} are particularly useful for comparing one or multiple representational spaces, they typically provide only a summary statistic of the degree of alignment and require explicit hypotheses and model comparisons to determine what it is about the representational space that drives human alignment. In contrast, other approaches have focused specifically on the interpretability of DNN representations \citep{erhan2009visualizing, zeiler2014visualizing, mahendran2015understanding,bau2017dissecting,  morcos2018importance,  nguyen2019understanding, bau2020understanding}, but either provide very specific local measures about DNN units or have limited direct comparability to human representations, as the same interpretability methods can typically not be applied to understand human mental representations. Our framework combines the strengths of the comparability gained from RSA and existing interpretability methods to understand image representations in DNNs. We applied common interpretability methods to show that our approach allows for detailed experimental testing and causal probing of DNN representations and behavior across diverse images. Yet, only the direct comparison to human representations revealed the diverging representational strategies of humans and DNNs and thus limitations of the visualization techniques we used \citep{geirhos2023dont}. 

Our results are consistent with previous work indicating that DNNs make use of  strategies that deviate from those used in humans \citep{geirhos2020shortcut, hermann2023foundations}. Beyond previously discovered biases, here we found a visual bias in DNNs that diverges from a semantic bias in humans for similarity judgments. Notably, even the highest layers in DNNs retained strong visual biases for solving the tasks they had been trained on, including image classification or linking images with text, which can both be described as semantic tasks with different degrees of richness. This visual strategy may, of course, reflect how our visual system solves core object recognition \citep{dicarlo2012does}. Indeed, it is an open question to what extent human core object recognition relies on a similar visual bias \citep{jagadeesh2022texture} and whether this bias is also found in anterior ventral-temporal cortex \citep{prince2024contrastive}, which is known to be involved in high-level object processing \citep{kanwisher2010functional} However, even if humans used a primarily visual strategy for solving core object recognition, our findings would demonstrate a significant limitation of DNNs in capturing human mental representations as measured with similarity judgments, despite similar representational geometries \citep{mur2013human}.

Interestingly, CLIP, a more predictive model of human cortical visual processing \citep{conwell2022can, wang2023better}, retained a visual bias despite training on semantic image descriptions, showing that classification objective alone is not sufficient for explaining visual bias in DNNs. At the same time, the visual bias in CLIP was smaller (see Supplementary Fig. \ref{fig:rsa-model-comps}b), indicating that better models of high-level visual processing may also be models with a larger semantic bias and pointing towards potential strategies for improving their alignment with humans, which may involve multimodal pretraining or larger, more diverse datasets \citep{wang2023better}. Future work would benefit from a systematic comparison of different DNNs to identify what factors determine visual bias and their alignment with human brain and behavioral data.

While these results indicate that studying core dimensions of DNN representations can improve our understanding of the factors required to identify better models of human mental representations, it has also been demonstrated recently that aligning DNNs with human representations can improve DNN robustness and performance at out-of-distribution tasks \citep{muttenthaler2023improving, sundaram2024does}. This work highlights that identifying visual bias may be useful not only for understanding representational and behavioral differences between humans and DNNs, but also for guiding future work determining the gaps in human-AI alignment and identifying adjustments in architecture, training needed to reduce this bias \citep{sucholutsky2023getting}. Further work is needed to clarify the role of task instructions in human-AI alignment across diverse tasks and instruction \citep{dwivedi2019representation}.

The framework introduced in this work can be expanded in several ways. Future work could use this approach to explore what factors make DNNs similar or different from one another. A comprehensive analysis of various DNN architectures, objectives, or datasets \citep{muttenthaler2022human, muttenthaler2023improving, conwell2022can} could uncover the factors underlying representational alignment, and extension to other stimuli, tasks, and domains, including brain recordings. Together, this framework promises a more comprehensive understanding of the relationship between human and AI representations, providing the potential to identify better candidate models of human cognition and behavior and more human-aligned artificial cognitive systems.

%TC:ignore
\section{Methods}\label{sec:methods}

\subsection{Triplet odd-one-out task}

In the triplet odd-one-out task, participants are presented with three objects and must choose the one that is least similar to the others, i.e. the odd-one-out. We define a dataset $\mathcal{D} \coloneqq \left\{\left(\{i_s, j_s, k_s\}, \{a_s, b_s\}\right)\right\}_{s=1}^{n}$ where \(n\) is the total number of triplets and \(\{i_s, j_s, k_s\}\) is a set of three unique objects, with \(\{a_s, b_s\}\) being the pair among them determined as most similar. We used a dataset of human responses collected by \citet{hebart2020revealing} to learn an embedding of human object concepts.  In addition, we simulated the triplet choices from a DNN. For the DNN, we simulated these choices by computing the dot product of the penultimate layer activation $\vz_{i} \in \mathbb{R}_{+}$ after applying the ReLU function, where $S_{ij} = \vz_i^\top \vz_j$. The most similar pair \( \{a_s, b_s\} \) was then identified by the largest dot-product: 

\begin{equation}
\{a_s, b_s\} = \argmax_{(x_s,y_s) \in \{(i_s,j_s), (i_s,k_s), (j_s,k_s)\}} \{\vz_{x_s}^\top \vz_{y_s}\}.
\end{equation}

Using this approach, we sampled triplet odd-one-out choices for a total of 20 million triplets for the DNN.

\subsection{Embedding optimization and pruning}\label{sec:embedding-optim-and-pruning}

\noindent{\bf{Optimization}.} Let $\mW \in \mathbb{R}^{m \times p}$ denote a randomly initialized embedding matrix, where $p=150$ is the initial embedding dimensionality. To learn interpretable concept embeddings, we used VICE, an approximate Bayesian inference approach \citep{muttenthaler2022vice}. VICE performs mean-field variational inference to approximate the posterior distribution $p(\mW | \mathcal{D})$ with a variational distribution, $q_{\theta}(\mW)$, where $q_{\theta} \in \mathcal{Q}$.

VICE imposes sparsity on the embeddings using a spike-and-slab Gaussian mixture prior to update the variational parameters $\theta$. This prior encourages shrinkage towards zero, with the spike approximating a Dirac delta function at zero (responsible for sparsity) and the slab modeled as a wide Gaussian distribution (determining non-zero values). Therefore, it is a sparsity-inducing prior and can be interpreted as a Bayesian version of the Elastic Net \citep{zou2005regularization}. The optimization objective minimizes the KL divergence between the posterior and the approximate distribution:

\begin{equation}
     \argmin_\theta~\mathbb{E}_{q_\theta\left(\mathbf{W}\right)}\left[\frac{1}{n}\log q_\theta\left(\mW\right) - \log p\left(\mW)\right) - \frac{1}{n}\sum_{s=1}^{n} \log p\left(\{a_{s},b_{s}\}|\{i_s, j_s, k_s\}, \mW\right)\right],
\end{equation}

where the left term represents the complexity loss and the right term is the data log-likelihood.

\noindent{\bf{Pruning}.} Since the variational parameters are composed of two matrices, one for the mean and one for the variance, $\theta = \{\mu, \sigma\}$, we can use the mean representation $\vmu_{i}$ as the final embedding for an object $i$. Imposing sparsity and positivity constraints improves the interpretability of our embeddings, ensuring that each dimension meaningfully represents distinct object properties. While sparsity is guaranteed via the spike-and-slab prior, we enforced non-negativity by applying a ReLU function to our final embedding matrix, thereby guaranteeing that $\mW \in \mathbb{R}_{+}^{m \times p}$. Note that this is done both during optimization and at inference time. We used the same procedure as in \citep{muttenthaler2022vice} for determining the optimal number of dimensions. Specifically, we initialized our model with $p=150$ dimensions and reduced the dimensionality iteratively by pruning dimensions based on their probability of exceeding a threshold set for sparsity:

\begin{equation}
    \text{Prune if } \Pr(w_{ij} > 0) < 0.05 \text{ for fewer than 5 objects},   
\end{equation}

where \( w_{ij} \) is the weight associated with object \(i\) and dimension \(j\). Training stopped either when the number of dimensions remained unchanged for 500 epochs or when the embedding was optimized for a maximum of 1000 epochs.

\subsection{Embedding reproducibility and selection}\label{sec:emb-reproducibility}

We assessed reproducibility across 32 model runs with different seeds using a split-half reliability test. We chose the split-half reliability test for its effectiveness in evaluating the consistency of our model's performance across different subsets of data, ensuring robustness. We partitioned the objects into two disjoint sets using odd and even masks. For each model run and every dimension in an embedding, we identified the dimension that is most highly correlated among all other models by using the odd-mask. Using the even mask, we correlated this highest match with the corresponding dimension. This process generated a sampling distribution of Pearson’s $r$ coefficients for all model seeds. We subsequently Fisher $z$-transformed the Pearson’s $r$ sampling distribution. The average $z$-transformed reliability score for each model run was obtained by taking the mean of these $z$-scores. Inverting this average provides an average Pearson’s $r$ reliability score (see Supplementary Information \ref{sec:reproducibility}). For our final model and all subsequent analyses, we selected the embedding with the highest average reproducibility across all dimensions.

\subsection{Labeling dimensions and construction of word clouds}\label{sec:labeling-dimensions}

We assigned labels to the human embedding by pairing each dimension with its highest correlating counterpart from \citet{hebart2020revealing}. These dimensions were derived from the same behavioral data, but using a non-Bayesian variant of our method. We then used the human-generated labels that were previously collected for these dimensions, without allowing for repeats. 

For the DNN, we labeled dimensions using human judgments. This allowed us to capture a broad and nuanced understanding of each dimension’s characteristics. To collect human judgments, we asked 12 laboratory participants (6 male, 6 female; mean age, 29.08; s.d. 3.09; range 25-35) to label each DNN dimension. Participants were presented with a $5 \times 6$ grid of images, with each row representing a decreasing percentile of importance for that specific dimension. The top row contained the most important images, and the following rows included images within the 8th, 16th, 24th, and 32nd percentiles. Participants were asked to provide up to five labels that they thought best described each dimension. Word clouds showing the provided object labels were weighted by the frequencies of occurrence, and the top 6 labels were visualized. Due to computer crashes during data acquisition, three participants had incomplete data (32\%, 80\% and 93\%).

Study participation was voluntary, and participants were not remunerated for their participation. This study was conducted in accordance with the Declaration of Helsinki and was approved by the local ethics committee of the Medical Faculty of the University Medical Center Leipzig (157/20-ek). 

\subsection{Dimension Ratings}

Two independent experts rated the dimensions according to two questions. The first question asked whether the dimensions were primarily visual-perceptual, semantic-conceptual, a mix of both, or whether their nature was unclear. For the second question, they rated the dimensions according to whether they reflected a single concept, several concepts, or were not interpretable. Overall, both raters agreed agreed 81.86\% of the time for question 1 and 90.00\% of the time for question 2. Response ambiguity was resolved by a third rater (see Supplementary Information \ref{sec:modelcomp}, \ref{sec:layercomp}, \ref{sec:interpretability}). All raters were part of the laboratory but were blind to whether the dimensions were model- or human-generated.

\subsection{Convolutional Embeddings}

We additionally learned embeddings from early (convolutional block 1), middle (convolutional block 3) and late (convolution block 5) convolutional layers of VGG-16. For this, we applied global average pooling to the spatial dimensions of the feature maps and then sampled triplets from the averaged 1d representations.

\subsection{Dimension value maximization}

To visualize the learned object dimensions, we used an activation maximization technique with a pretrained StyleGAN XL generator $\mathcal{G}$ \citep{sauer2022stylegan}. Our approach combines sampling with gradient-based optimization to generate images that maximize specific dimension values in our embedding space.

\textbf{Initial sampling.} We started by sampling a set of $N = 100,000$ concatenated noise vectors $\vv_i \in \mathbb{R}^d$, where $d$ is the dimensionality of the StyleGAN XL latent space. For each noise vector, we generated an image $\vx_i = \mathcal{G}(\vv_i)$ and predicted its embedding $\hat{\vy}_i \in \mathbb{R}^p$ using our pipeline, where $p$ is the number of dimensions in our embedding space.

For a given dimension $j$, we selected the top $k$ images that yielded the highest values for $\hat{y}_{ij}$, the $j$-th component of $\hat{\vy_i}$. These images served as starting points for our optimization process.

\textbf{Gradient-based optimization.} To refine these initial images, we performed gradient-based optimization in the latent space of StyleGAN XL. Our objective function $\mathcal{L_\text{AM}}$ balances two goals: increasing the absolute value of the embedding for dimension $j$ and concentrating probability mass towards dimension $j$. Formally, we define $\mathcal{L}_\text{AM}$ as:

\begin{equation}
\mathcal{L}_\text{AM}(\vv_i) = -\alpha \cdot \hat{y}_{ij} - \beta \cdot \log p\left(\hat{y}_{ij} \mid \vz_i\right),
\end{equation}

where $\vz_i = f(\mathcal{G}(\vv_i))$ denotes the penultimate features extracted from the generated image using the pretrained VGG-16 classifier $f$. The term on the left, referred to as the \textit{dimension size reward}, contributes to increasing the absolute value $\hat{y}{ij}$ for the object dimension $j$. The term on the right, referred to as the \textit{dimension specificity reward}, concentrates probability mass towards a dimension without necessarily increasing its absolute value. The balance between these two objectives is controlled by the scalars $\alpha$ and $\beta$. The objective $\mathcal{L}_\text{AM}$ was minimized using vanilla stochastic gradient descent. Importantly, only the latent code vector $\vv_i$ was updated, while keeping the parameters of $\mathcal{G}$, the VGG-16 classifier $f$, and the embedding model fixed.

This optimization process was performed for each of the top $k$ images selected in the initial sampling phase. The resulting optimized images provide visual representations that maximally activate specific dimensions in our learned embedding space, offering insights into the semantic content captured by each dimension.

\subsection{Highlighting image properties}

To highlight image regions driving individual DNN dimensions, we used Grad-CAM. For each image, we performed a forward pass to obtain an image embedding and computed gradients using a backward pass. We next aggregated the gradients across all feature maps in that layer to compute an average gradient, yielding a two-dimensional dimension importance map.

\subsection{RSA analyses}

We used RSA to compare the structure of our learned embeddings with human judgments and DNN features. This analysis was conducted in three stages: human RSA, DNN RSA, and a comparative analysis between human and DNN representations.

\textbf{Human RSA.} We reconstructed a similarity matrix from our learned embedding. Given a set of objects $\mathcal{O} = {o_1, \dots, o_m}$, we computed the similarity $S_{ij}$ between each pair of objects $(o_i, o_j)$ using the softmax function:

\begin{equation}
S_{ij} = \frac{1}{|\mathcal{O} \setminus \{o_i, o_j\}|} \sum_{k \in \mathcal{O} \setminus \{o_i, o_j\}} \frac{\exp\left(\vy_i^\top \vy_j\right)}{\exp\left(\vy_i^\top \vy_j\right) + \exp\left(\vy_i^\top \vy_k\right) + \exp\left(\vy_j^\top \vy_k\right)}
\end{equation}

where $\vy_i$ is the embedding of object $o_i$, and the softmax function returns the probability of $o_i$ being more similar to $o_k$ than $o_j$. To evaluate the explained variance, we used a subset of 48 objects for which a fully sampled similarity matrix and associated noise ceilings were available from previous work \citep{hebart2020revealing}. We then computed the Pearson correlation between our predicted RSM and the ground truth RSM for these 48 objects.

\textbf{DNN RSA.} We followed a similar procedure, reconstructing the RSM from our learned embedding of the DNN features. We then correlated this reconstructed RSM with the ground-truth RSM derived from the original DNN features used to sample our behavioral judgments.

\textbf{Comparative Analysis.} To compare human and DNN representations, we conducted two analyses. First, we performed a pairwise comparison by matching each human dimension with its most correlated DNN dimension. This was done both with and without replacement, allowing us to assess the degree of alignment between human and DNN representational spaces. Second, we performed a cumulative RSA to determine the number of DNN dimensions needed to accurately reflect the patterns in the human similarity matrix. We took the same ranking of DNN dimensions used for the pairwise RSA, starting with the highest correlating dimension. We then progressively added one DNN dimension at a time to a growing subset. After each addition, we reconstructed the RSM from this subset and correlated both the human RSM and the cumulative DNN RSM. This step-by-step process allowed us to observe how the inclusion of each additional DNN dimension contributed to explaining the variance in the human RSM.

% Added this for Review and all OSF repostories
\section*{Data Availability}
The images used in this study are from the THINGS object concept and image database, available at the OSF repository (\href{https://osf.io/jum2f/}{https://osf.io/jum2f}). All result files pertaining to this study are made publicly available via a separate OSF repository (\href{https://osf.io/nva43/}{https://osf.io/nva43/}).

\section*{Code availability}
A Python implementation of all experiments presented in this paper is publicly available at \href{https://github.com/florianmahner/object-dimensions/}{https://github.com/florianmahner/object-dimensions/} \citep{mahner2025object}. 

\section*{Acknowledgments}
FPM, LM and MNH acknowledge support by a Max Planck Research Group grant of the Max Planck Society awarded to MNH. MNH acknowledges support by the ERC Starting Grant COREDIM (ERC-StG-2021-101039712) and the Hessian Ministry of Higher Education, Science, Research and Art (LOEWE Start Professorship and Excellence Program “The Adaptive Mind”). U.G. acknowledges support from the project Dutch Brain Interface Initiative (DBI2) with project number 024.005.022 of the research program Gravitation which is (partly) financed by the Dutch Research Council (NWO). LM acknowledges support by the German Federal Ministry of Education and Research (BMBF) for the Berlin Institute for the Foundations of Learning and Data (BIFOLD) (01IS18037A) and for the grants BIFOLD22B and BIFOLD23B. This study used the high-performance from the Raven and Cobra Linux clusters at the Max Planck Computing \& Data Facility (MPCDF), Garching, Germany \href{https://www.mpcdf.mpg.de/services/supercomputing/}{https://www.mpcdf.mpg.de/services/supercomputing/}. The funders had no role in study design, data collection and analysis, decision to publish or preparation of the manuscript. 

\section*{Author Contributions}
Conceptualization: FPM, LM, MNH; Funding acquisition: MNH; Software: FPM, LM; Supervision: MNH; Visualization: FPM, LM; Writing – original draft: FPM, LM; MNH Writing – final manuscript: FPM, LM, UG, MNH.

\section*{Competing interests}
There are no financial and non-financial competing interests.

\bibliographystyle{plainnat}
% \bibliography{bibliography} 

\clearpage

\appendix
\renewcommand{\thesection}{\Alph{section}}
\setcounter{section}{0}
\setcounter{figure}{0}
\renewcommand{\thefigure}{\arabic{figure}}

\section{Dimension ratings and RSA across models}\label{sec:modelcomp}

\begin{figure}[!htb]
    \centering
    \includegraphics[width=\textwidth]{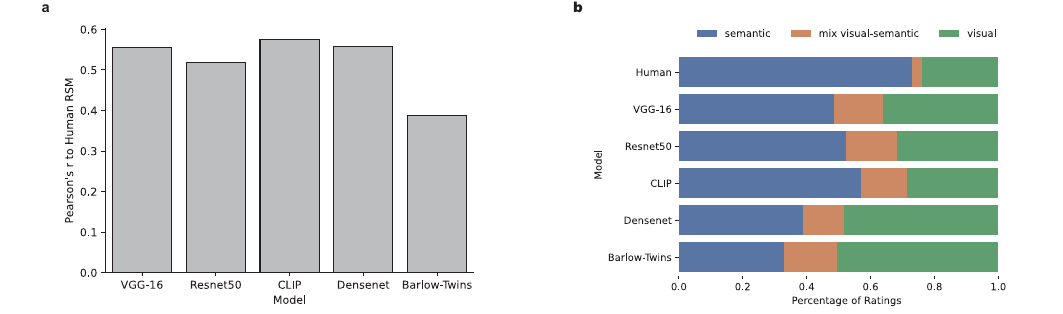}
    \caption{\textbf{Dimension ratings and representational similarity across models}. \textbf{a}, VGG-16 does not perform poorly when compared to other models, including Resnet50, DenseNet, CLIP, and BarlowTwins-Resnet50. \textbf{b}, The visual bias identified in VGG-16 is also evident across these other architectures, demonstrating consistent differences between human and DNN dimensions.}
    \label{fig:rsa-model-comps}
\end{figure}

 We used VGG-16 due to its common use in computational cognitive neuroscience. To validate that VGG-16 is a suitable choice for the comparison to humans, we conducted RSA analyses with  various other DNN models that differ in training diets, objective functions, and architecture. Using \texttt{thingsvision} \citep{thingsvision}, we similarly extracted the penultimate features of these models and learned representational embeddings based on simulated triplet choices. Each embedding was then compared to the human-derived one using RSA. In Supplementary Fig. \ref{fig:rsa-model-comps}a we can see that VGG-16 does not perform poorly compared to other architectures,  which suggests that it is a suitable choice for our analyses. Additionally, we assessed the visual bias in these architectures by having human raters categorize each dimension's dominant visual property as visual, semantic, a mixture of both, or unclear. This reveals that the visual bias we find for VGG-16 also replicates across different DNNs (Supplementary Fig. \ref{fig:rsa-model-comps}b).

\section{Dimension ratings and RSA across layers}\label{sec:layercomp}

\begin{figure}[!htb]
\centering
\includegraphics[width=\textwidth]{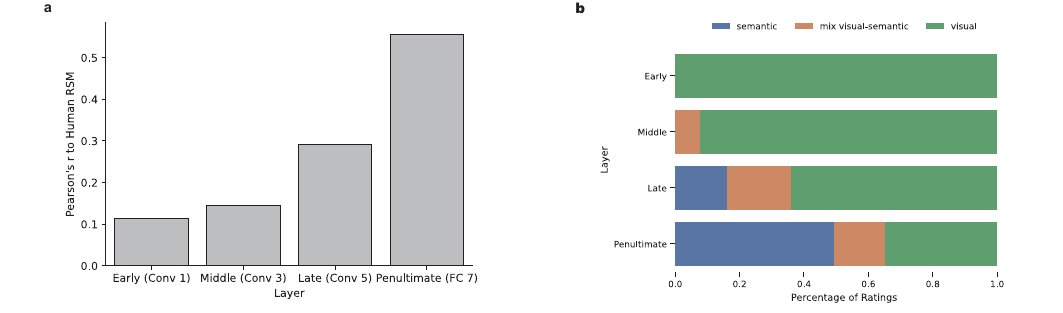}
\caption{\textbf{Dimension ratings and representational similarity across different VGG-16 layers}. We compared dimensions learned from features extracted from early, middle, late and penultimate layers of VGG-16. \textbf{a}, The embedding learned from the penultimate layer representations has the largest representational alignment to human behavior. \textbf{b}, The visual bias was strongest in early layers of VGG-16 and semantic information is added in later layers. The fraction of semantic dimensions compared to visual dimensions is largest in the penultimate embedding.}
\label{fig:early-late}
\end{figure}

We demonstrate that the visual bias was present throughout the network with a gradient from early to late layers, with early representations showing the largest visual bias and penultimate representations the smallest. In addition, the penultimate embedding also exhibited the strongest human alignment (Supplementary Fig. \ref{fig:early-late}).

\section{Human ratings of dimension interpretability}\label{sec:interpretability}

\begin{figure}[!htb]
    \centering
    \includegraphics[width=\textwidth]{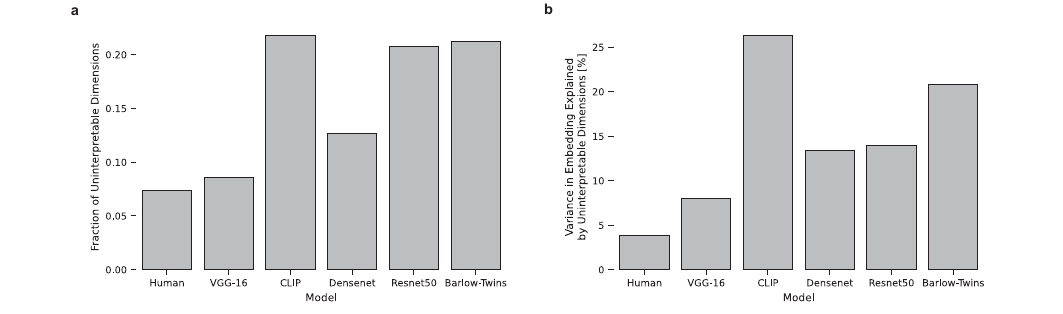}
    \caption{\textbf{Dimension ratings for different DNN models}. \textbf{a}, Percentage of interpretable dimensions as rated by human observers. Across all DNN models, the human embedding has the smallest percentage of uninterpretable dimensions. \textbf{b}, Variance explained by uninterpretable dimensions. For this, we weighted the uninterpretable dimensions with their importance as given by the numeric value of that dimension. Compared to humans, all uninterpretable DNN dimensions explain more variance in their embedding.}
\label{fig:interpretability}
\end{figure}

 To gain an understanding of how meaningfully interpretable DNN dimensions were, we additionally asked the human experts to rate the interpretability of the dimensions across all five DNN architectures. Across DNNs, the number of interpretable dimensions was consistently lower than that found in humans (see Supplementary Fig. \ref{fig:interpretability}a), and uninterpretable dimensions generally had a higher overall importance for odd-one-out choices as indicated by the sum of their weight across images (embedding variance explained by uninterpretable dimensions: 3.83\% Humans, 8.02\% VGG-16, Supplementary Fig. \ref{fig:interpretability}b). Taken together, despite the decent global alignment between human and DNN representations and numerous interpretable visual and semantic dimensions, these results demonstrate largely different strategies used by humans and DNNs for object processing, with DNNs using more visual properties and exhibiting a stronger mix between visual and semantic information than humans, who primarily rely on semantic properties. This visual bias in DNNs is accompanied by an overall reduced interpretability of dimensions as compared to humans, demonstrating that sparse and non-negative embeddings do not necessarily result in interpretable dimensions and indicating another potential deviation of DNNs to the way humans represent visual stimuli.

\section{Task validation} \label{sec:task-validation}

We investigated to what degree the instruction to focus on the entire image would affect triplet choice behavior. To this end, we sampled 300 random triplets of the 1,854 objects 80 times each, once with an instruction to focus on the object (as before) and repeated the same experiment with an instruction to focus on the entire image. We ensured that we selected non-overlapping sets of participants. All participants provided informed consent, and the study was approved by the local ethics committee of the Medical Faculty of the University Medical Center Leipzig (157/20-ek). Due to a coding error, participant age was not recorded. A total of 713 participants took part in this study. Of them, 91 were removed due to too fast responses, leaving 276 for the object focus group (166 male, 109 female, 1 other) and 346 for the image focus group (186 male, 159 female, 1 other). We analyzed participants’ consistency in their triplet responses for each task separately and tested to what degree comparing between the tasks would lead to a reduction in this consistency. The results showed numerically higher noise ceilings for the image focus than the object focus (noise ceiling image focus: $r = 0.8955$, noise ceiling object focus: $r = 0.8743$), indicating slightly higher data quality. Importantly, the consistency between object focus and image focus triplet choices was indistinguishable from the object focus noise ceiling (consistency: $r = 0.8720$, $p > 0.05$, based on 1000 bootstrap samples). Together, this demonstrates that the instruction likely did not affect how participants carried out the task.

\section{Semantic embedding validation}\label{sec:sem-validation}

\begin{figure}[!htb]
    \centering
    \includegraphics[width=\textwidth]{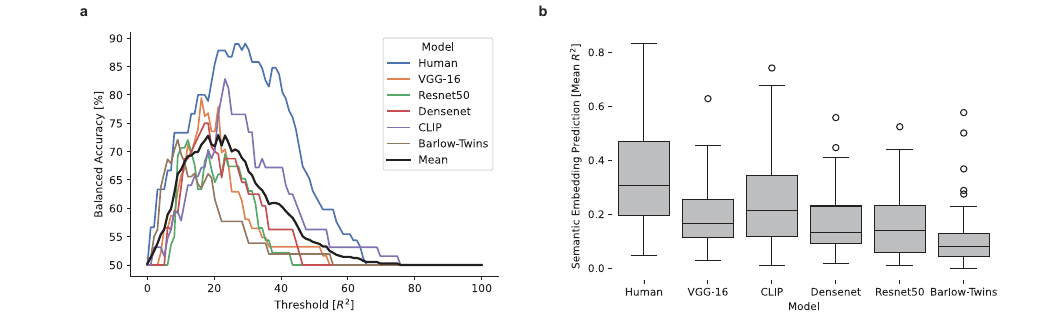}
    \caption{ \textbf{Validating human ratings via semantic text embeddings}. \textbf{a}, We plot the balanced accuracy (average of sensitivity and specificity) of labeling dimensions as visual or semantic based on the predictions of the semantic text embedding for varying thresholds. We can see that using the semantic text embedding the human ratings can be explained for all models. All models peak between $R^2$ = [20\%, 30\%]. This shows that human raters have a systematic visual bias that equally applied to human and DNN models. \textbf{b}, For each model, we plot the explained variance in predicting each dimension using the semantic text embedding. The box plots show the median (center), 25th and 75th percentiles (box bounds) and whiskers extend to the minima and maxima within 1.5 times the interquartile range.}
    \label{fig:sem-validation}
\end{figure}

To address the degree to which dimensions could be explained as being semantic, in addition to the human ratings, we used a semantic embedding based on nouns corresponding to the objects and predicted each dimension’s values along the images from this embedding. We reasoned that a semantic embedding that had been trained only on text would be better at predicting dimensions that are semantic in nature than visual dimensions, and in turn, the predictivity should tell us about the degree to which a dimension can be classified as visual or semantic. In this context, high predictivity from a semantic embedding would indicate a semantic dimension, while low predictivity would indicate a non-semantic, likely visual dimension. To test this, we filtered the DNN embedding to include only the 1,854 objects viewed by human participants and excluded mixed and unclear dimensions from both the human and DNN embeddings to remove ambiguity and make results more comparable. We then used cross-validated ridge regression to predict each human and DNN dimension from a 300-dimensional semantic embedding matrix \citep{pilehvar2016conflated} across the 1,854 objects evaluated by humans, leading to one $R^2$ value per dimension. Since sparser dimensions will lead to lower overall prediction accuracy irrespective of whether a dimension is visual or semantic, we adjusted the $R^2$ across all dimensions within each model by fitting a linear regression model to the predicted $R^2$ scores across dimension positions and removing the linear component. Please note that, while this procedure improved overall prediction of whether a dimension qualifies as semantic or visual, this did not affect the overall pattern of results. The resulting adjusted $R^2$ scores represent each dimension's semantic importance independent of its position, allowing for fairer comparison between early and late dimensions. This analysis revealed that, for all models, including human judgments, we can reliably distinguish between predominantly visual and semantic dimensions, with up to 90\% balanced accuracy for humans (Supplementary Fig. \ref{fig:sem-validation}a). In addition, we found a very similar semantic bias across humans and neural networks using this measure (Supplementary Fig. \ref{fig:sem-validation}b). Together, these results support the human ratings and suggest that both human and DNN dimensions can be classified reliably as either visual or semantic, with similar biases observed across models, highlighting that the effects we found cannot be explained solely by a human rating bias.

\clearpage
\section{Causal Image Manipulations}\label{sec:causal}

\begin{figure}[!htb]
    \centering
    \includegraphics[width=\textwidth]{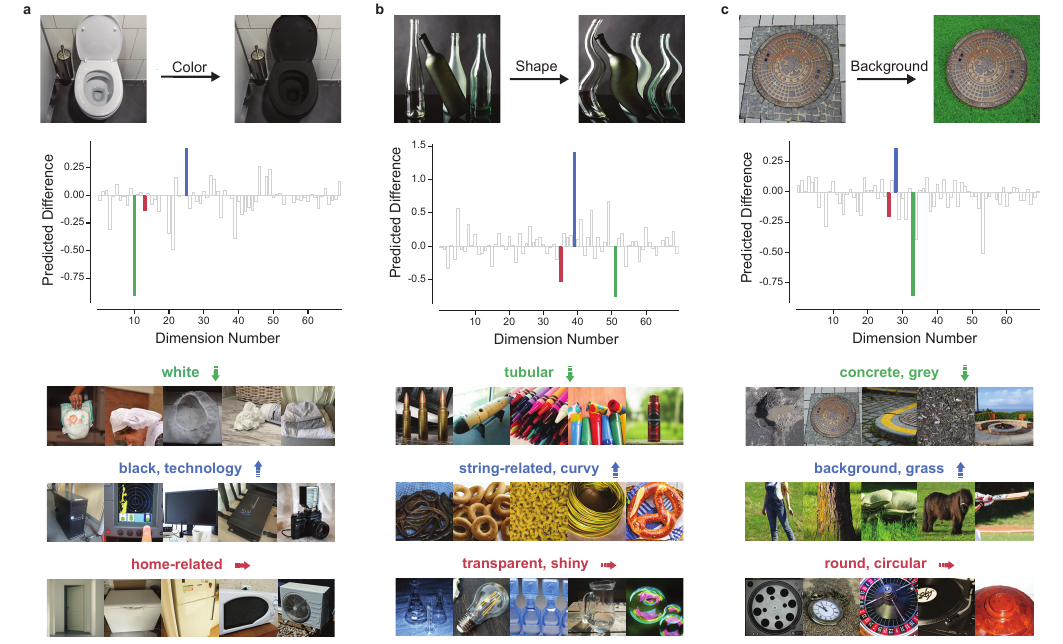}
    \caption{\textbf{Causal manipulation of unique image properties}. We compared the predicted dimension values between the original and causally manipulated images using our interpretability pipeline, revealing how these manipulations specifically affected various dimensions within our embedding space. The arrows indicate whether the activation level of a dimension increases, decreases, or remains relatively unchanged due to the manipulation. \textbf{a}, Altering the color of a toilet from white to black, \textbf{b}, Modifying the shape of a set of bottles to be more curved, \textbf{c}, Changing the background in an image containing a manhole. Note that the displayed images reflect only images with a public domain license and not the full image set \citep{stoinski2023thingsplus}.}
    \label{fig:causal}
\end{figure}

\clearpage

\section{Reproducibility}\label{sec:reproducibility}
\begin{figure}[!htb]
    \centering
    \includegraphics[width=\textwidth]{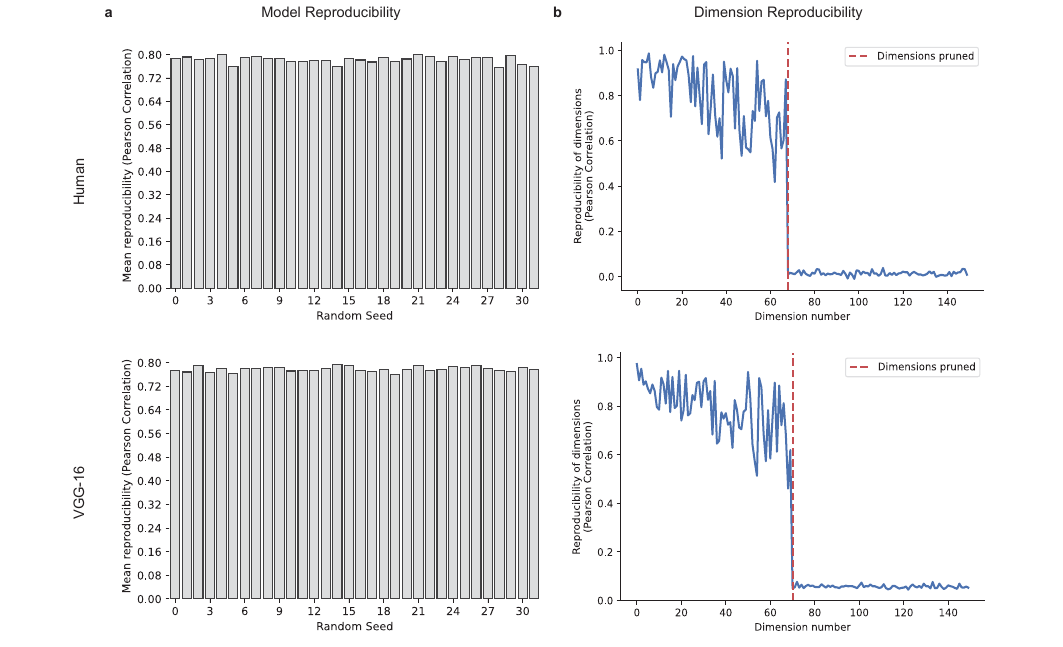}
    \caption{\textbf{Model reproducibility across different random initializations in humans and the DNN.} \textbf{a}, Reproducibility across model runs was evaluated using a split-half reliability test (see Methods: \nameref{sec:emb-reproducibility}). The model with the highest average reproducibility was selected for subsequent experiments. \textbf{b}, For this model, we present a visualization of its dimensional reproducibility compared to other models and dimensions. The red line indicates the number of dimensions retained in the final model as determined by the VICE criteria.}
    \label{fig:reproducibility}
\end{figure}
%TC:endignore

\end{document}